\documentclass[11pt,a4paper]{article}
\usepackage{authblk}
\usepackage[hyperref]{emnlp2020}
\usepackage{times}
\usepackage{paralist}
\usepackage{latexsym}

\usepackage{graphicx, multirow}
\usepackage{todonotes}
\usepackage{enumitem}

\usepackage{microtype}

\aclfinalcopy

\newcommand{\intex}{\textsc{InterExp}}
\newcommand{\summ}{\sigma}
\newcommand{\sess}{\Sigma}
\newcommand{\sect}{\S}

\title{Evaluating Interactive Summarization: an Expansion-Based Framework}

\author[1]{\bf Ori Shapira}
\author[2]{\bf Ramakanth Pasunuru}
\author[3]{\bf Hadar Ronen}
\author[2]{\\ \bf Mohit Bansal}
\author[1]{\bf Yael Amsterdamer}
\author[1]{\bf Ido Dagan}
{
\makeatletter
\renewcommand\AB@affilsepx{~~~~~~ \protect\Affilfont} \makeatother
\affil[1]{Bar-Ilan University}
\affil[2]{UNC Chapel Hill}
\affil[3]{Peres Academic Center}
}
\affil[  ]{} % skip line of affiliations
\affil[  ]{\tt \{obspp18, hadarg\}@gmail.com}
\affil[  ]{\tt \{ram, mbansal\}@cs.unc.edu}
\affil[  ]{\tt \{amstery, dagan\}@cs.biu.ac.il}

\date{}

\begin{document}
\maketitle
\begin{abstract}
Allowing users to interact with multi-document summarizers is a promising direction towards improving and customizing summary results.
Different ideas for interactive summarization have been proposed in previous work but these solutions are highly divergent and incomparable. In this paper, 
we develop an end-to-end evaluation framework for \emph{expansion-based} interactive summarization, which considers the accumulating information along an interactive session. Our framework includes a procedure of collecting real user sessions and evaluation measures relying on standards, but adapted to reflect interaction.
All of our solutions are intended to be released publicly as a benchmark, allowing comparison of future developments in interactive summarization.
We demonstrate the use of our framework by evaluating and comparing baseline implementations that we developed for this purpose, which will serve as part of our benchmark.
Our extensive experimentation and analysis of these systems motivate our design choices and support the viability of our framework.
\end{abstract}

\section{Introduction}
\label{sec_introduction}
Exploratory search~\citep{marchionini2006exploratorySearch} is a process where users interact with a system to navigate through a large knowledge repository, and acquire information according to their needs. We consider exploration in the context of multi-document summarization (MDS), where the ability to interact with a system to affect summary contents may
help users improve upon automatically generated summaries, without having to resort back to the full texts. For example, users can identify missing information in a summary and accordingly ask the system to expand on specific contents.

A key gap in the development and adoption of interactive summarization solutions is the lack of a benchmark for reliable and meaningful comparison between different systems, similarly to benchmarks for static (non-interactive) summarization \citep[e.g.,][]{nist2014DUCWebsite}. Some research studies previously proposed interesting solutions for interactive or customizable summarization \citep[e.g.,][]{baumel2014qcfs, christensen2014hierarchicalSumm, handler2017rookie, leuski2003ineats, lin2010interactiveMMR, shapira2017ias, yan2011personalizedSumm}. However, these solutions are distinct, with proprietary evaluations that do not admit comparison.

In this paper we develop an end-to-end evaluation framework for interactive summarization systems, allowing to compare system performance. Our framework supports a general notion  of \textit{expansion-based} interactive summarization (\intex{}), where the textual summary gradually expands in response to user interaction, influenced by the user's interests (\sect\ref{sec_definition}).
See Figure \ref{fig_qfseSystem} for an example of a user interaction with an \intex{} system, gradually expanding an initial, automatically generated summary by posing textual queries.
To ensure our framework is sound, we developed it in multiple cycles accompanied by user studies and extensive crowdsourcing experimentation. Our main contributions may be summarized as follows. 

\emph{Evaluation measures.} We propose a few evaluation measures for \intex{}~systems,
which build upon a combination of established notions in automated summarization and interactive systems, and specifically enable utilization of available MDS datasets as input. Our measures are aggregated over multiple interactive sessions and document sets to obtain an overall reliable system evaluation. In contrast to static summarization, our measures apply to the different steps along the interaction to reflect the progress of information acquirement rather than just its final result.
See \sect\ref{sec_evalOfSessions}.

\emph{Implementation.} To demonstrate the viability of the evaluation framework, we develop two baseline systems for extractive \textit{query-focused summary expansion} (QFSE), namely, where users pose textual queries to the system and
are responded with relevant sentences from the source texts.
To allow for interactivity, much emphasis is placed on low latency algorithms, an issue that is often overlooked in MDS, and NLP in general. See \sect\ref{sec_implementations}.

\emph{Session collection procedure.} To allow collecting real user sessions adequate for \intex{}~system comparison, we describe a
controlled crowdsourcing procedure
that minimizes noise and effect of subjective preferences in collected data.
This procedure makes our evaluation framework accessible and relatively scalable for researchers interested in pursuing the task, when compared to other tasks involving interactive systems.
See \sect\ref{sec_sessionCollection}.

\emph{Experimental study.} Finally, we demonstrate the use of our evaluation framework over our two baselines, using the DUC 2006~\cite{dang2006DUCoverview}
dataset and controlled crowdsourcing. 
Analysis shows favorable results in terms of overall system quality and internal consistency between sessions, users, and different evaluation measures, indicating that our solutions may serve as a promising benchmark for \intex{}~system evaluation. See \sect\ref{sec_experiments}.

\section{Background}
\label{sec_background}
Traditional MDS has been researched extensively over the past few decades \citep[e.g.][]{goldstein2000extractiveMDS, radev2004centroidMDS, haghighi2009exploringMDS, yin2015neuralMDS}. It encompasses variants of \emph{query-focused summarization} \citep{dang2005DUCoverview}, orienting the output summary around a given query \citep[e.g.][]{daume2006bayesianQFS, zhao2009qfs, cao2016attsum, feigenblat2017unsupervisedQFS, baumel2018QFS}, and \emph{incremental update summarization} \citep{dang2008TACoverview}, generating a summary of a set of documents with the assumption of prior knowledge on an earlier document set \citep[e.g.][]{li2008pnr2, wang2010updateSumm, mccreadie2014incremental, zopf2016sequentialUpdateSumm}.

The \emph{query-chain focused} summarization task \citep{baumel2014qcfs} integrates the query-focused and incremental update setups. A chain of queries yields a sequence of short summaries, each refraining from repeating content. Although somewhat interactive in nature, the task's evaluation is based solely on pre-defined sequences of queries with a respective reference summary per iteration that disregards previous outputs by the system.
The hierarchical summarization work by \citet{christensen2014hierarchicalSumm} presents a preassembled summary with several levels of detail, allowing a user to drill down to information of interest.
The interactive summarization system by \citet{shapira2017ias} provides expansions on an initial summary on the information-unit level, also pre-generated.
The two last works do not provide an evaluation measurement that is comparable to other interactive systems or that factor in the information variation due to interaction.

Exploratory search \citep{marchionini2006exploratorySearch, white2009exploratory} attempts to, more formally, address the need for converting big data to knowledge by imposing cooperation of algorithms and humans. For example, interactive information retrieval (IIR) \citep{ingwersen1992iir} focuses on fine-tuning relevant document retrieval interactively, and complex interactive question answering (ciQA) \citep{kelly2007overviewTrec} involves interacting with a system to generate a passage that answers a complex question.

Evaluation is a major challenge in dealing with any of the aforementioned interactive tasks~\citep{white2008evaluatingExpS, palagi2017explSearchModels, hendahewa2017evaluatingExpSeTrails}. Firstly, real users must use the system being evaluated, requiring an (often expert) user study that highly increases the cost and complexity of the evaluation process.
Then, output assessments can be hazy due to varying user behavior on the system. We address these issues below.

\section{Task Definition}
\label{sec_definition}
The input to a session of an \intex{}~system~$S$ is a \emph{multi-document set}~$D$, typically on a certain topic. $S$~processes these documents and first outputs a short initial summary $\summ_0$ to provide the gist of the input source texts. $\summ_0$ is an \emph{informative body of text}, as opposed to keyword lists, visual timelines or any other non-summarized or non-textual forms.
To obtain additional information, the user interacts with the system, e.g., by pressing a button or submitting a query. For each user-guided interaction $q_i$ the system computes a corresponding response $r_i$ that provides further summarized textual information (like $\summ_0$) on the document's topic. 
Consequently, these responses may be viewed as \emph{expansions} of $\summ_0$, gradually building up a more complete summary of the topic: $\summ_0\cup r_1$, then $\summ_0\cup r_1\cup r_2$, etc. To be practical, the computation time of the initial summary as well as of interaction responses is required to meet \emph{interactivity standards} \citep{Anderson2020WebsiteLoad, attig2017latencyGuidelines}, e.g., less than half a second for interaction response computation.

\section{System Evaluation}
\label{sec_evalOfSessions}
For compared systems ${S_1, S_2, ..., S_m}$, we require at least $u$ sessions of distinct users interacting with $S_i$ on each test document set $D\in\{D_1, D_2, ..., D_n\}$.
Assuming such sessions, we next define automatic and manual evaluation measures according to the task definition above, and defer details on adequate session collection to \sect\ref{sec_sessionCollection}.

\subsection{Automatic Measures}
\label{sec_automaticmeasures}

Formally, we model a session as a tuple $\sess=(S,D,\summ_0,[q_1,\dots,q_{|\sess|}],[r_1,\dots,r_{|\sess|}])$ where $S$ is an \intex{} system, $D$ is a set of documents, $q_i$-s and $r_i$-s are interaction representations and responses respectively, and $|\sess|$ is the number of interactions posed during the session. $\sess$ defines a sequence of incrementally expanding \emph{snapshots} $[\summ_0,\summ_1,\dots,\summ_{|\sess|}]$ where $\summ_i=\summ_0\cup \bigcup_{j=1}^i r_j$ is the union of accumulative (summarized) information presented to the user after~$i$ interactions. Each snapshot may thus be viewed as a static summary.

Our first challenge is thus obtaining comparable scores at different length summaries (snapshots).
While we wish to make use of static MDS benchmarks, these provide reference summaries at a \textit{single} length with the goal of generating a summary at a \textit{similar} length.
To address this challenge, we leverage a finding by \citet{shapira2018multiLenRouge} showing that a reference summary of a single length can be used to relatively compare varying length summaries on a topic.

Based on this observation we now define three indicators for system performance, first over a single session and then aggregated over all sessions of a system.

\begin{figure}[t]
    \centering
    \includegraphics[width=1.0\columnwidth]{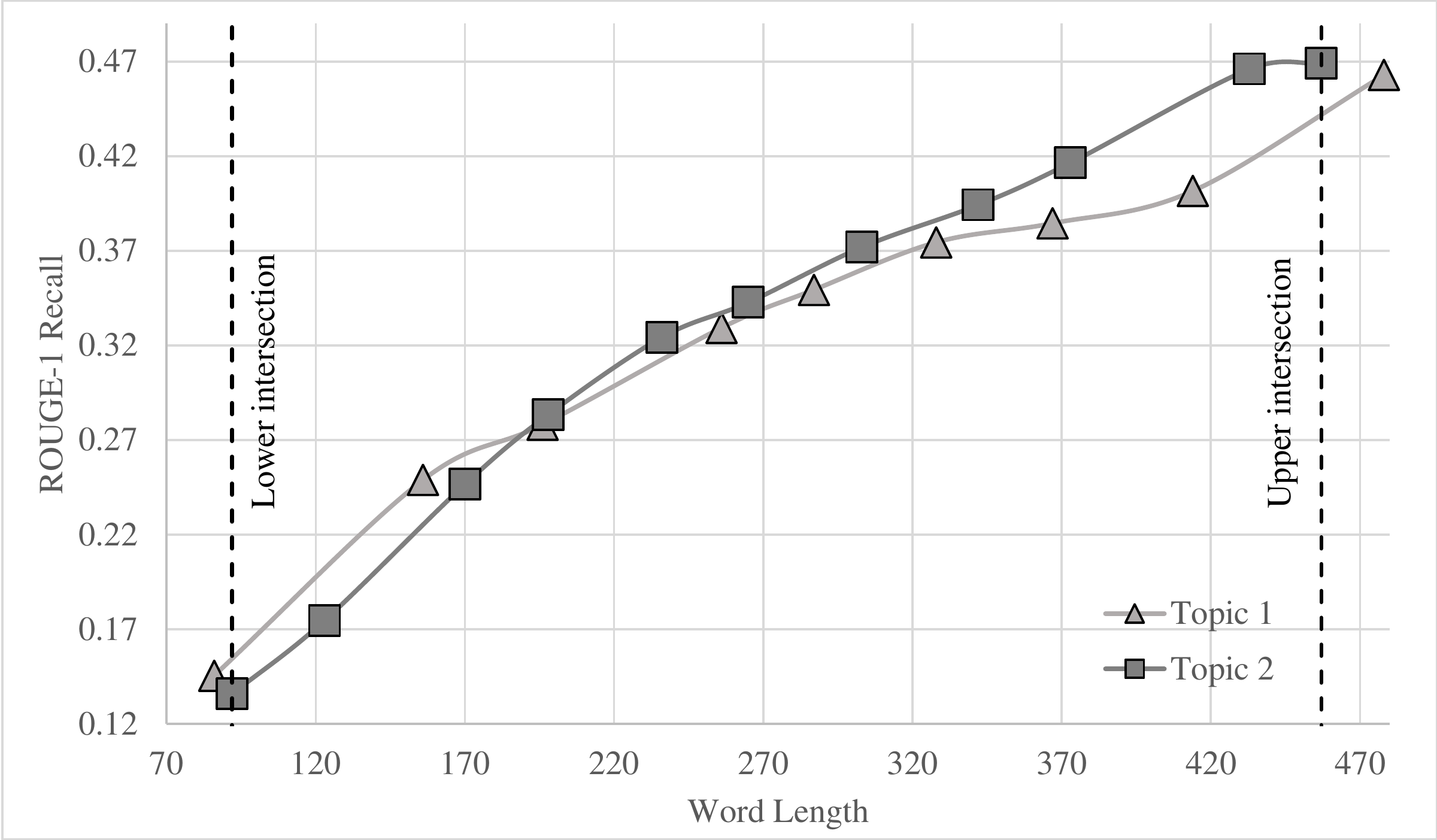}
    \caption{Example recall-curves of two sessions on an \intex{}~system. Points plotted per interaction snapshot within a session. Range of intersection between observed summary lengths is bounded by dashed lines.}
    \label{fig_recallCurves}
\end{figure}

\paragraph*{Per-session indicators.} (1) To illustrate the gradual information gain along a session we adopt a \emph{recall-by-length curve}~\citep{kelly2007overviewTrec, lin2007recallCurve}, see for example Figure~\ref{fig_recallCurves}. The curve's x-axis is the snapshot word-length, chosen as the dominant factor affecting quality, as opposed to number of queries or interaction time, which are not necessarily comparable between sessions.
The y-axis is a content recall score, such as ROUGE-recall \citep{lin2004rouge}.\footnote{Any content recall measure can be employed as long as it is consistent, including, e.g., manual mechanisms like Pyramid or nugget-style scoring \citep{nenkova2004pyramid, lin2006nuggetPyr}.} For a session $\sess$ with snapshots ${\summ_0, \summ_1,...,\summ_{|\sess|}}$, each snapshot, $\summ_i$, with word length $x_i$ and content recall score $y_i$ is plotted on the graph at point $(x_i, y_i)$.

(2) We consider the \emph{area under the recall-curve} (AUC). Intuitively, it is desirable for an \intex{} system to generate more salient information earlier: assuming salient information is more relevant to users, this property will allow users to terminate the interaction sooner, as soon as their information needs are met. Accordingly, AUC is higher when content is more relevant and is retrieved earlier. AUC is defined between start and end x-values, fixed for comparable measurement (see Figure~\ref{fig_recallCurves}), and the y-value scores are interpolated at these points when the relevant curves do not have a snapshot at the specific length.

(3) We consider the \emph{Score@Length} metric that reports a score, such as ROUGE $F_1$, at pre-specified word-lengths.
This metric enables fair comparison to static summaries at the specified lengths.
The inverse Length@Score measure is also examined, and detailed further in the supplementary material.

\paragraph*{Aggregated indicators.} 
Our final performance indicators at the system level are computed from the respective session indicators, as follows.
\begin{itemize}[nosep,leftmargin=1em,labelwidth=*,align=left]
\item The \emph{average recall curve}, illustrating overall gradual information gain, is computed from individual session recall-at-length curves by interpolating y-values at constant x-value increments and averaging correspondingly. E.g., Figure \ref{fig_averageCurve}. 
\item $[P.1]$ is the average AUC computed from the individual session AUCs by first averaging per topic and then averaging the results over all topics, to give equal weight to each topic.
\item $[P.2]$ is the average Score@Length computed similarly to average AUC from individual session Score@Lengths.
\end{itemize}

The evident advantages of our proposed evaluation framework are: (1) our automatic indicators are absolute and comparable from one session/system to another; (2) our evaluation framework fundamentally extends upon prevailing static summarization evaluation practices and utilizes existing MDS dataset reference summaries.

\subsection{Human Ratings}
\label{sec_humanratings}
Our evaluation framework allows doubly leveraging the involvement of human users by asking them to rate different system aspects during the session.
We suggest the rating layout listed as follows, with each measurement being scored on a 1-to-5 scale.
\begin{itemize}[nosep,leftmargin=1em,labelwidth=*,align=left]
\item $[R.1]$ After reading the initial summary, the user rates how informative it is for the given topic.
This somewhat resembles the DUC manual content rating \citep{dang2006DUCoverview}.
\item $[R.2]$ To measure the information gain throughout the session, the user rates how much useful information each interaction's response adds.
As this rating is scored per interaction, the session average is utilized as a measure for overall ability to expose interesting information. 
\item $[R.3]$ After the session, the user rates how well, overall, the responses responded to the requests throughout the session.
\item $[R.4]$ The user rates the two statements constructing the UMUX-Lite \citep{lewis2013umuxlite} questionnaire, measuring perceived system usability: $[R.4a]$ the system's capabilities meet the requirements 
and $[R.4b]$ the system is easy to use. 
The UMUX-Lite score is a function of these two scores (although $[R.4a]$ and $[R.4b]$ are separately useful) and shows high correlation to the popular, and longer, SUS questionnaire \citep{brooke1996sus}, thus offering a cheaper alternative.
\end{itemize}

Similarly to our automatic measures, these ratings are collected separately per session and then averaged, first per topic and then over all topics, to obtain a final system score allowing to compare between systems.

\section{Baseline Algorithm Implementations}
\label{sec_implementations}
According to the \intex{} task definition, we implemented two baseline systems, allowing to gather user sessions and demonstrate the utility of our evaluation framework. As mentioned above, the systems and all relevant data and results will be released publicly\footnote{\url{github.com/OriShapira/InterExp}} to serve as baselines and a benchmark for the \intex{} task.

The systems that we constructed are both based on what we term \emph{query-focused summary expansion} (QFSE), namely, user interactions are textual queries and the responses aim to maximize relevance to the query, similarly to query-focused static summarization, while refraining from repeating previously presented contents, somewhat similarly to update summarization.
The implementations support queries from free-text, highlights and system suggestions.
To simplify the analysis, both systems are extractive and use the same GUI, and in particular, the same interaction modes. Both systems produce bullet-style texts, where discourse and anaphora are not considered.

\subsection{Algorithm Variants}

Each of our implementations consists of three main components, as follows.

\paragraph{Initial summary.}
We first consider the generation of the initial summary $\summ_0$. Our experimentation with some classic \textit{and} modern MDS implementations have indicated that most do not meet the interactivity response-time requirements, and hence we provide two implementations for this component based on standard extractive MDS methods.

The first algorithm, denoted $I^{\mathrm{CL}}$, is clustering-based with ideas inspired by \citet{rossiello2017centroidBasedSumm} and \citet{hong2014wordImportance}. All sentences in the document set are separately assigned a representation by
averaging the 300-dimensional word2vec (w2v) embeddings \citep{mikolov2013distributed} within each sentence.
Each vector's dimension is reduced to 20 with PCA \citep{wold1987pca}, and then sentence vectors are clustered to 30 components with k-means \citep{macqueen1967kmeans}. We then order clusters by size, and select a representing sentence starting from the largest cluster, and continuing to following clusters, until the summary word-limit has reached. A cluster is skipped if its representing sentence is too similar (cosine similarity of 0.95) to previously selected sentences. The sentence selected to represent a cluster is the one whose words are on average most frequent within the full document set. The hyperparameters were lightly adjusted by computing ROUGE scores of some outputs against reference summaries and ensuring fast processing. On a standard server, the implementation can generate a summary of 25 news articles from a common MDS dataset in a few seconds (2 to 10 seconds).

The second algorithm is TextRank \citep{mihalcea2004textrank}, denoted $I^{\mathrm{TR}}$.
It is slightly slower than $I^{\mathrm{CL}}$, but runs within approximately the same time range.

\paragraph{Query response generation.}
This component computes similarity of a given query to all sentences not yet presented in previous iterations, and outputs a few of the best matches. Process-time is up to a few hundred milliseconds.

The first variant, $Q^{\mathrm{SEM}}$, utilizes the semantic (w2v-based) sentence representations prepared in the initialization process.
It computes the cosine similarity between the query's representation and all unused source sentences. In MMR-style \citep{goldstein2000mmr}, the sentences most similar to the query 
which pass a dissimilarity threshold
from already selected sentences are selected.
An MMR dissimilarity of $\leq 0.05$ gave the best results (on simulated sessions).

The second variant, $Q^{\mathrm{LEX}}$, additionally considers lexical similarity by scoring the similarity of a query and a sentence as the product of the w2v similarity with three ROUGE-precision scores. I.e., for query $q$ and sentence $s$, $\mathrm{sim}(q,s)=(\mathrm{cosine}(w2v(q),w2v(s))+1)*(\mathrm{R1}_p(q,s)+1)*(\mathrm{R2}_p(q,s)+1)*(\mathrm{RL}_p(q,s)+1)$, where $\mathrm{R1}_p$, $\mathrm{R2}_p$ and $\mathrm{RL}_p$ are ROUGE-1, ROUGE-2 and ROUGE-L precision respectively. The highest scoring unused sentences are output. Compared to the first variant, this method yields lexically-stricter search results.

\paragraph{Suggested queries.}
Our QFSE systems support an interaction mode enabling information expansion by clicking a system-suggested query from a list.
The first approach for preparing the list, $\mathrm{Sug}^{\mathrm{FREQ}}$, is selecting the most frequent bigrams and trigrams within the source documents, disregarding stop words. A trigram is preferred when it contains a bigram with the same frequency.

The second approach, $\mathrm{Sug}^{\mathrm{TR}}$, utilizes the top-ranked phrases extracted from the TextRank algorithm (as part of the summarization procedure).

\paragraph{Overall systems.}
For the purpose of applying our evaluation framework, we picked
two combinations of the above three components. (Assessing additional combinations is out of the scope of this paper.) The first combination, denoted System $S_1$, is comprised of $I^{\mathrm{CL}}$, $Q^{\mathrm{SEM}}$ and $\mathrm{Sug}^{\mathrm{FREQ}}$.
The second combination, System $S_2$, consists of $I^{\mathrm{TR}}$, $Q^{\mathrm{LEX}}$ and $\mathrm{Sug}^{\mathrm{TR}}$.\footnote{\label{footnote_suppMaterial}See appendix for further details.}

While developing these algorithms, we also experimented with Sentence-BERT \citep{reimers2019sentenceBert} representations in $I^{\mathrm{CL}}$, and with BERTScore \citep{zhang2019bertscore} as a similarity score between the query and each potential sentence. These variants did not improve outputs, and, more importantly, were impractical due to their considerably higher latency.

\subsection{Web Application}
\label{sec_webApp}
We developed a front-end web application for session collection with real users. The application communicates with a QFSE system in the back-end -- in our case $S_1$ or $S_2$.

The initial version of the application was assessed via a small-scale user study of~10 users. We used an SUS questionnaire \citep{brooke1996sus} and applied the think-aloud protocol \citep{lewis1982thinkaloud} to get feedback on how to improve the application and observe the users' reaction to the system.

\begin{figure}[t]
    \centering
    \includegraphics[width=1.0\columnwidth]{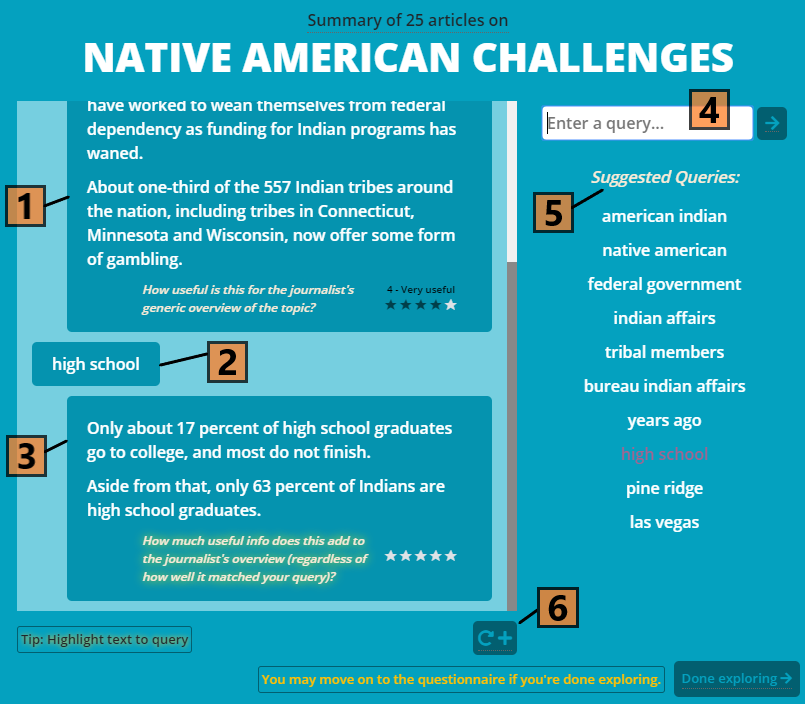}
    \caption{The QFSE web application: [1] Initial summary with user rating $[R.1]$; [2] user query; [3] system response with user rating used in $[R.2]$; [4] query box; [5] clickable system suggested queries; [6] button to repeat last query.}
    \label{fig_qfseSystem}
\end{figure}

Figure \ref{fig_qfseSystem} displays a screenshot of the improved web
application, on the topic ``Native American Challenges''. The figure shows the initial summary [1], a user query [2] and the corresponding response [3] in the summary pane.
[4] features the query box, where users can input a query as free text (user generated query), highlight text from the presented summary as a query (summary-inspired query, pasted automatically to the query box) or choose from the list of suggested queries (system hint), shown in [5]. The last query can also be repeated via a button [6],
to obtain additional information on the same query.

\section{Session Collection}
\label{sec_sessionCollection}
The evaluation of interactive systems requires \emph{real user sessions}. Several of our own cycles of session collection uncovered multiple user-related challenges, in-line with previous work on user task design
\citep{christmann2019conversational, roit2019controlled, zuccon2013crowdsourcingInteractions}.
Recruited users may make undue interactions due to insincere work or experimental behavior, yielding noisy sessions that do not reflect realistic system performance. Additionally, without a clear and objective informational goal, a user will interact with the system according to subjective interests, yielding sessions that are incomparable by objective measures.

\paragraph{Controlled crowdsourcing method.}
Employing experts to use the system in a user study is not scalable or feasible for many researchers, making crowdsourcing an appealing and comparatively inexpensive alternative. We carefully designed\textsuperscript{\ref{footnote_suppMaterial}} a three-stage controlled crowdsourcing protocol that mitigates the above mentioned challenges for improved session quality, while filtering out users with low engagement or lacking relevant skills.

The first stage is a three-question \textit{trap task} whose aim is to efficiently filter out insincere workers and conversely discover workers with an ability to apprehend salient information within text. The second stage assigns \textit{practice tasks} that familiarize the users to the actual \intex{} system interface, to prevent experimentation in the evaluated sessions. Here, the users are also presented with a \emph{grounding use-case}, or `cover-story' as termed by \citet{borlund2003iirEvaluation}. The use-case states an objective common goal to follow in interacting with the system, to minimize the effect of subjective preferences. An example use-case to follow, which we applied in our experiments, is ``produce an informative summary draft text which a journalist could use to best produce an overview of the topic''. This use-case  is strongly emphasized during practice sessions, with additional integrated guidelines throughout the session.
Workers completing two practice assignments with predominantly relevant interactions are invited to continue on to the final stage.

The \textit{evaluation session collection} stage involves interacting with the evaluated system, for a minimum amount of time per session (e.g., 150 seconds in our experiments), to produce a summary on a topic in light of the same assigned use-case as in the practice stage. Each worker may perform at most one task per topic, and the overall goal is recording sufficiently many sessions per combination of system and topic, for sufficiently many topics -- e.g., in our experiments in \sect\ref{sec_experiments} we used~20 topics and required at least~3 sessions per system per topic.

\paragraph*{Wild versus controlled crowdsourcing.} We illustrate the benefit of the controlled crowdsourcing procedure described above by comparing its results with a ``wild'' crowdsourcing preliminary experiment.
The latter experiment applied basic filtering of users (workers with a $99\%$ approval rate and at least 1000 approved assignments on Amazon Mechanical Turk\footnote{\url{https://www.mturk.com}} (AMT)) and skipped straight to the evaluation session collection stage. A questionnaire was filled out by workers at the end of each session as a post-hoc quality control mechanism to hypothetically filter out insincere workers.

\begin{table}
    \centering
    \resizebox{\columnwidth}{!}{
    \begin{tabular}{l|c|c}
        \hline
        \textbf{Measure} & \textbf{Controlled} & \textbf{Wild} \\
        \hline
        \# interactions & 12.3 & 7.0 \\
        Approx.\ explore time & 250 sec. & 170 sec. \\
        \% suggested query & $36.2\%$ & $62.7\%$ \\
        \% free-text query & $25.3\%$ & $2.2\%$ \\
        \% $\Delta$AUC from lower bound & $+1.8\%$ & $-1.4\%$ \\
        \hline
    \end{tabular}}
    \caption{Qualitative measures of improved session collection through controlled crowdsourcing against sincere wild crowdsourcing. Values are computed per session and averaged over all sessions on System $S_1$.}
    \label{table_wild_vs_controlled}
\end{table}

Analysis of the collected sessions showed a substantial improvement in querying behavior in controlled over wild crowdsourcing --  the former scored higher than the latter on every evaluation metric.
Table \ref{table_wild_vs_controlled} presents some qualitative indications of this improvement: controlled users were more engaged (more iterations and more time exploring) and put more thought into their queries (more free-text queries and less suggested queries).
Notably, unlike uncontrolled crowd-workers, controlled workers were able to do better than a comparable fully automated baseline, supporting an \emph{interactivity hypothesis}: human interaction enables a summarization system to produce better outputs.
This is evident from the last table row, displaying the percent difference in ROUGE-1 AUC score from a lower bound simulated baseline (explained in \sect\ref{sec_simulatedBounds}), which is positive (better) for controlled sessions and negative (worse) for wild sessions.
Finally, the queries of controlled users almost exclusively adhered to the use-case and the many helpful comments from the workers indicated their attentiveness to the task.

\section{Experiments}
\label{sec_experiments}

We demonstrate the proposed evaluation framework by comparing the two QFSE systems $S_1$ and $S_2$ described in \sect\ref{sec_implementations}.

\subsection{Crowd Experimental Setup}
Following our procedure from \sect\ref{sec_sessionCollection}, we released the trap task in AMT on~10 topics, each with~39 assignments. 48~of~231 workers qualified for the second stage, out of which~25 accepted. 10~workers passed the training stage, out of which we recruited 8~highly qualified workers. For the third stage, we collected sessions for~20 topics from DUC 2006, on $S_1$ and $S_2$. Each worker could explore~10 different topics on each system, amounting to~160 possible sessions of which~153 were completed, and~3-4 sessions for each combination of topic and system. A minimum exploration time constraint of~150 seconds was set. Initial summaries were a minimum of~75 tokens (average about~85), and responses were two sentences long.

The full controlled crowdsourcing process took one author-work-week, and cost~\$370. In comparison, ``wild'' crowdsourcing described in \sect\ref{sec_sessionCollection} required a couple days' work and~\$240 (achieving, as shown above, inferior results), and running a non-crowdsourced user-study of the same magnitude would likely require more work time, and cost an estimated~\$480 (32~net hours of 16 workers at a commonly acceptable~\$15 hourly wage). Furthermore, the results of a user study would not necessarily provide higher quality results \citep{zuccon2013crowdsourcingInteractions}. To our judgement, the controlled crowdworkers are more suitable due to understanding the task \textit{before} choosing to complete it. 

\subsection{Simulated Bounds}
\label{sec_simulatedBounds}
In addition to real user experiments, we simulate each of our two systems on two scripted query lists, one for offering a lower bound case and one for an upper bound case. Here again, each simulation's initial summary has a minimum length limit of 75 words, and responses are two sentences long.

The first query list, $L^{\mathrm{Sug}}$, is constructed fully automatically: it consists of the top-10 ordered phrases in the system's suggested queries component per topic, i.e. $\mathrm{Sug}^{\mathrm{FREQ}}$ in $S_1$, and $\mathrm{Sug}^{\mathrm{TR}}$ in $S_2$. This mimics a ``lower bound'' user who adopts the simplest strategy, namely, clicking the suggested queries in order without using personal judgment even to choose among these queries.

The second list, $L^{\mathrm{Oracle}}$, consists of 10 randomly chosen crowdsourced summary content units (SCUs) \citep{shapira2019litepyramids} for each of the topics. Since the SCUs were extracted from the reference summaries of the corresponding topics, they mimic a user who searches for the exact information required to maximize similarity to the same reference summaries which we then evaluate against.
Hence, this ``oracle'' simulation serves as our upper bound.

\begin{figure}[t]
    \centering
    \includegraphics[width=1.0\columnwidth]{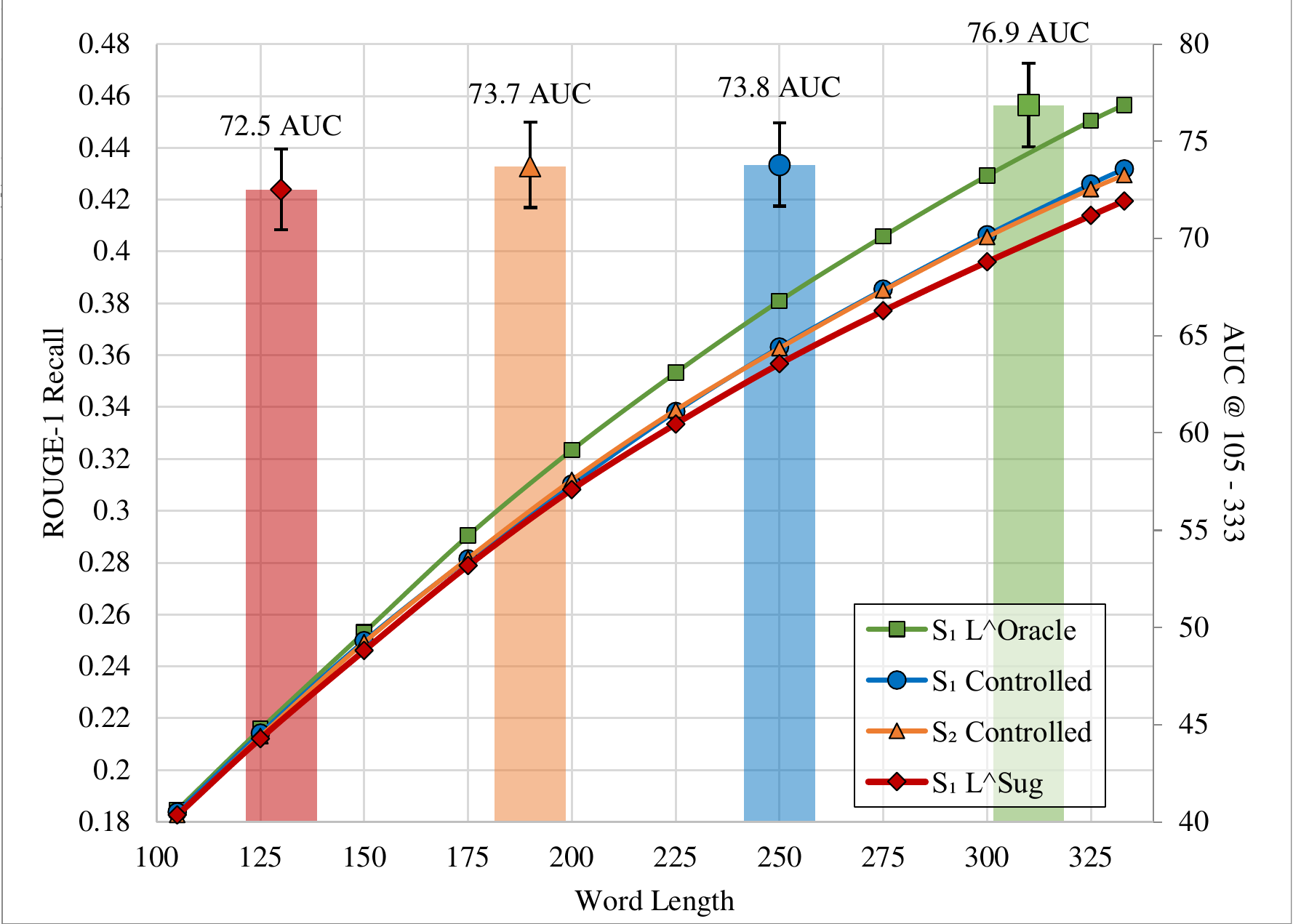}
    \caption{The average recall-curves, along with corresponding AUC scores (unrelated to the x-axis) and their confidence intervals ($\geq95\%$), of the upper and lower bound sessions and of user sessions of the two systems.}
    \label{fig_averageCurve}
\end{figure}

\subsection{Experimental Results}
Figure \ref{fig_averageCurve} presents the average recall-curves and corresponding $[P.1]$ averaged AUC scores
of the~$S_1$ bounds and of the user sessions on $S_1$ and $S_2$. AUC is computed between word-lengths~105 to~333 (the maximum intersection of all sessions) and the intervals at $95\%$ confidence (bootstrapped).
Table~\ref{table_scores} shows $[P.2]$ averaged ROUGE-1 based Score@Length.\footnote{Giving a comparative sense of these performances relative to recent costly state-of-the-art algorithms for static MDS, the latter (S@L 250) are similar to our Oracle upper bounds, while our user sessions performances correspond to (simpler) algorithms needed to achieve interactive response time.} Scores rank consistently on ROUGE-2, ROUGE-L and ROUGE-SU.\textsuperscript{\ref{footnote_suppMaterial}}

\begin{table}
    \centering
    \resizebox{\columnwidth}{!}{
    \begin{tabular}{l|c|c|c}
        \hline
        \textbf{Sessions} & \textbf{S@L 150} & \textbf{S@L 250} & \textbf{S@L 350} \\
        \hline
        $S_1$ $L^{Oracle}$ & $.328$ ($\pm .012$) & $.400$ ($\pm .010$) & $.414$ ($\pm .012$) \\
        $S_1$ Real & $.324$ ($\pm .010$) & $.382$ ($\pm .011$) & $.392$ ($\pm .012$) \\
        $S_1$ $L^{Sug}$ & $.319$ ($\pm .011$) & $.375$ ($\pm .012$) & $.382$ ($\pm .011$) \\
        \hline
        $S_2$ $L^{Oracle}$ & $.333$ ($\pm .011$) & $.402$ ($\pm .013$) & $.412$ ($\pm .014$) \\
        $S_2$ Real & $.321$ ($\pm .009$) & $.379$ ($\pm .013$) & $.388$ ($\pm .012$) \\
        $S_2$ $L^{Sug}$ & $.320$ ($\pm .011$) & $.374$ ($\pm .014$) & $.386$ ($\pm .013$) \\
        \hline
    \end{tabular}}
    \caption{ROUGE-1 $F_1$-based average Score@Length of simulated sessions vs.\ real user sessions. Scores at 350 words are approximate as few sessions were shorter. Scores rank consistently on ROUGE-2, -L and -SU. Intervals at $\geq 95\%$ confidence.}
    \label{table_scores}
\end{table}

First, it is evident from Figure~\ref{fig_averageCurve} and  Table~\ref{table_scores}  that the results on collected sessions \emph{indeed fall between the two bounds in all measures}. This demonstrates the effectiveness of interactive summarization, even when using relatively simple algorithms: the algorithm enables fast information processing of input texts, while users effectively direct the algorithm to 
salient areas.

Second, the scores of $S_1$ and $S_2$ are highly close, providing no significant insights when comparing these two systems, which is surprising due to their distinct implementations. We have manually reviewed the results and were convinced that they are consistent with intuition -- indeed, the systems happen to perform at similar quality overall. However, when assessing the systems' separate components and inspecting user-provided ratings, we gain awareness of some interesting distinctions.

Table \ref{table_systemMetrics} shows a trend of consistency between ROUGE scores on each separate component and the ratings provided by the users. The initial summaries' ROUGE-1 $F_1$ scores are computed against the reference summaries, with a slight advantage for $I^{\mathrm{CL}}$ in $S_1$ over $I^{\mathrm{TR}}$ in $S_2$ -- similar to the users' initial summary ratings.
For the query-response component, we compute the average ROUGE $F_1$ score of the independent responses to the queries in $L^{\mathrm{Oracle}}$ (a jointly consistent list of queries), against the reference summaries.
Again, user ratings reflect a similar trend that $Q^{\mathrm{LEX}}$ of $S_2$ slightly outscores $Q^{\mathrm{SEM}}$ of $S_1$. Overall we see that $S_1$ provides a better initial summary while $S_2$ handles queries better. Also, users tend to be more satisfied by $S_2$, possibly due to its ability to respond better to queries. This claim is also evident from the positive correlation between $[R.3]$ and $[R.4a]$, $r=0.68, p<0.001$ in $S_1$ and $r=0.63, p<0.001$ in $S_2$. In terms of absolute UMUX-Lite scores $[R.4]$, 68~is considered average, and above~80 is considered excellent, meaning both $S_1$ and $S_2$ got high usability scores.\textsuperscript{\ref{footnote_suppMaterial}}

An additional analysis finds a positive correlation between per-iteration response $[R.2]$ scores and the relative per-iteration increase in ROUGE recall (e.g.\ for ROUGE-1 $r=0.36, p<0.001$ in $S_1$ and $r=0.33, p<0.001$ in $S_2$), hinting at the credibility of correlation between human ratings and relative increase in ROUGE within sessions.

To conclude, more systems are needed for a more conclusive appraisal of the full evaluation framework, regardless of the accidental similarity between our baselines. Yet, our findings are already favourable in terms of internal consistency of measures and soundness of the computed scores.

\begin{table}
    \centering
    \resizebox{\columnwidth}{!}{
    \begin{tabular}{c|l|c|c}
        \hline
        & \textbf{Metric} & \textbf{$S_1$} & \textbf{$S_2$} \\
        \hline
        \hline
        \parbox[t]{1mm}{\multirow{2}{*}{\rotatebox[origin=c]{90}{Initial}}} & Initial summ.\ ROUGE-1 & \textbf{0.232} & 0.225 \\
        & $[R.1]$ Initial summ.\ rating & \textbf{3.89} (0.98) & 3.71 (1.0) \\
        \hline
        \parbox[t]{1mm}{\multirow{3}{*}{\rotatebox[origin=c]{90}{Query}}} & Que-Resp $L^{Oracle}$ ROUGE-1 & 0.156 & \textbf{0.161} \\
        & $[R.2]$ Avg.\ resp.\ rating & 3.17 (1.32) & \textbf{3.35} (1.28) \\
        & $[R.3]$ Query responsiveness & 3.61 (1.02) & \textbf{3.83} (1.03) \\
        \hline
        \parbox[t]{1mm}{\multirow{3}{*}{\rotatebox[origin=c]{90}{Overall}}} & $[R.4a]$ System effectiveness & 3.81 (1.0) & \textbf{4.05} (0.80) \\
        & $[R.4a]$ System ease of use & 4.51 (0.71) & \textbf{4.63} (0.62) \\
        & $[R.4]$ System UMUX-Lite & 74.2 (12.5) & \textbf{77.1} (10.3) \\
        \hline
    \end{tabular}}
    \caption{Average and (StD) scores of metrics comparing $S_1$ and $S_2$. Users appear to be more satisfied with $S_2$ overall, likely due to the query response component.}
    \label{table_systemMetrics}
\end{table}

\section{Discussion and Conclusion}
\label{sec_discussion}
We formally define the \intex{} task as interactive summarization by iterations of user-guided information expansions.
Compared to previous interactive tasks, advantages of our evaluation framework and controlled crowdsourcing procedure are apparent in their ability to compare between systems with relative ease. The implemented baseline systems and utilized data, to be released publicly, provide the means to advance relevant research and allow further experimentation, which are also required to substantiate the credibility and reveal the full potential of our evaluation framework.

In future work, it is worthwhile to separately assess the effectiveness of individual interaction modes, including ones incorporated in our implementation and others. Within our expansion-based framework, we can consider additional measures of textual consistency, coherence, and relevance of responses to queries. We may also test additional approaches for summarization: e.g., \emph{abstractive} summarization for flexible synthetic summary generation, requiring further evaluation of factuality and truthfulness. Beyond our framework, that targets objective quality, \intex{}~systems should also be evaluated according to their compatibility with personalized, subjective use.

\section*{Acknowledgments}

We would like to thank Guiseppe Carenini for his helpful advice.
This work was supported in part by the German Research Foundation through the German-Israeli Project Cooperation (DIP, grants DA 1600/1-1 and GU 798/17-1); by the BIU Center for Research in Applied Cryptography and Cyber Security in conjunction with the Israel National Cyber Bureau in the Prime Minister's Office; by the Israel Science Foundation (grants 1157/16 and 1951/17); by a grant from the Israel Ministry of Science and Technology; by the NSF-CAREER Award \#1846185; and by a Microsoft PhD Fellowship.

\bibliography{bibliography}

\begin{thebibliography}{57}
\expandafter\ifx\csname natexlab\endcsname\relax\def\natexlab#1{#1}\fi

\bibitem[{Anderson(2020)}]{Anderson2020WebsiteLoad}
Shaun Anderson. 2020.
\newblock \href
  {https://www.hobo-web.co.uk/your-website-design-should-load-in-4-seconds}
  {How fast should a website load?}
\newblock
  https://www.hobo-web.co.uk/your-website-design-should-load-in-4-seconds.
\newblock Accessed: 2020-05-19.

\bibitem[{Attig et~al.(2017)Attig, Rauh, Franke, and
  Krems}]{attig2017latencyGuidelines}
Christiane Attig, Nadine Rauh, Thomas Franke, and Josef~F Krems. 2017.
\newblock System latency guidelines then and now--is zero latency really
  considered necessary?
\newblock In \emph{International Conference on Engineering Psychology and
  Cognitive Ergonomics}, pages 3--14. Springer.

\bibitem[{Baumel et~al.(2014)Baumel, Cohen, and Elhadad}]{baumel2014qcfs}
Tal Baumel, Raphael Cohen, and Michael Elhadad. 2014.
\newblock \href {https://doi.org/10.3115/v1/P14-1086} {Query-chain focused
  summarization}.
\newblock In \emph{Proceedings of the 52nd Annual Meeting of the Association
  for Computational Linguistics (Volume 1: Long Papers)}, pages 913--922,
  Baltimore, Maryland. Association for Computational Linguistics.

\bibitem[{Baumel et~al.(2018)Baumel, Eyal, and Elhadad}]{baumel2018QFS}
Tal Baumel, Matan Eyal, and Michael Elhadad. 2018.
\newblock Query focused abstractive summarization: Incorporating query
  relevance, multi-document coverage, and summary length constraints into
  seq2seq models.
\newblock \emph{arXiv preprint arXiv:1801.07704}.

\bibitem[{Borlund(2003)}]{borlund2003iirEvaluation}
Pia Borlund. 2003.
\newblock The iir evaluation model: a framework for evaluation of interactive
  information retrieval systems.
\newblock \emph{Information research}, 8(3):8--3.

\bibitem[{Brooke(1996)}]{brooke1996sus}
John Brooke. 1996.
\newblock Sus-a quick and dirty usability scale.
\newblock \emph{Usability evaluation in industry}, 189(194):4--7.

\bibitem[{Cao et~al.(2016)Cao, Li, Li, Wei, and Li}]{cao2016attsum}
Ziqiang Cao, Wenjie Li, Sujian Li, Furu Wei, and Yanran Li. 2016.
\newblock Attsum: Joint learning of focusing and summarization with neural
  attention.
\newblock In \emph{Proceedings of COLING 2016, the 26th International
  Conference on Computational Linguistics: Technical Papers}, pages 547--556.

\bibitem[{Christensen et~al.(2014)Christensen, Soderland, Bansal, and
  {Mausam}}]{christensen2014hierarchicalSumm}
Janara Christensen, Stephen Soderland, Gagan Bansal, and {Mausam}. 2014.
\newblock \href {https://doi.org/10.3115/v1/P14-1085} {Hierarchical
  summarization: Scaling up multi-document summarization}.
\newblock In \emph{Proceedings of the 52nd Annual Meeting of the Association
  for Computational Linguistics (Volume 1: Long Papers)}, pages 902--912,
  Baltimore, Maryland. Association for Computational Linguistics.

\bibitem[{Christmann et~al.(2019)Christmann, Saha~Roy, Abujabal, Singh, and
  Weikum}]{christmann2019conversational}
Philipp Christmann, Rishiraj Saha~Roy, Abdalghani Abujabal, Jyotsna Singh, and
  Gerhard Weikum. 2019.
\newblock Look before you hop: Conversational question answering over knowledge
  graphs using judicious context expansion.
\newblock In \emph{Proceedings of the 28th ACM International Conference on
  Information and Knowledge Management}, pages 729--738.

\bibitem[{Dang(2005)}]{dang2005DUCoverview}
Hoa~Trang Dang. 2005.
\newblock Overview of duc 2005.
\newblock In \emph{Proceedings of the document understanding conference},
  volume 2005, pages 1--12.

\bibitem[{Dang(2006)}]{dang2006DUCoverview}
Hoa~Trang Dang. 2006.
\newblock Overview of duc 2006.
\newblock In \emph{Document Understanding Conference}.

\bibitem[{Dang and Owczarzak(2008)}]{dang2008TACoverview}
Hoa~Trang Dang and Karolina Owczarzak. 2008.
\newblock Overview of the tac 2008 update summarization task.
\newblock In \emph{TAC}.

\bibitem[{Daum{\'e}~III and Marcu(2006)}]{daume2006bayesianQFS}
Hal Daum{\'e}~III and Daniel Marcu. 2006.
\newblock Bayesian query-focused summarization.
\newblock In \emph{Proceedings of the 21st International Conference on
  Computational Linguistics and the 44th annual meeting of the Association for
  Computational Linguistics}, pages 305--312. Association for Computational
  Linguistics.

\bibitem[{Feigenblat et~al.(2017)Feigenblat, Roitman, Boni, and
  Konopnicki}]{feigenblat2017unsupervisedQFS}
Guy Feigenblat, Haggai Roitman, Odellia Boni, and David Konopnicki. 2017.
\newblock Unsupervised query-focused multi-document summarization using the
  cross entropy method.
\newblock In \emph{Proceedings of the 40th International ACM SIGIR Conference
  on Research and Development in Information Retrieval}, pages 961--964.

\bibitem[{Goldstein et~al.(2000{\natexlab{a}})Goldstein, Mittal, Carbonell, and
  Callan}]{goldstein2000mmr}
Jade Goldstein, Vibhu Mittal, Jaime Carbonell, and Jamie Callan.
  2000{\natexlab{a}}.
\newblock Creating and evaluating multi-document sentence extract summaries.
\newblock In \emph{Proceedings of the ninth international conference on
  Information and knowledge management}, pages 165--172.

\bibitem[{Goldstein et~al.(2000{\natexlab{b}})Goldstein, Mittal, Carbonell, and
  Kantrowitz}]{goldstein2000extractiveMDS}
Jade Goldstein, Vibhu Mittal, Jaime Carbonell, and Mark Kantrowitz.
  2000{\natexlab{b}}.
\newblock Multi-document summarization by sentence extraction.
\newblock In \emph{Proceedings of the 2000 NAACL-ANLP Workshop on Automatic
  summarization}, pages 40--48. Association for Computational Linguistics.

\bibitem[{Haghighi and Vanderwende(2009)}]{haghighi2009exploringMDS}
Aria Haghighi and Lucy Vanderwende. 2009.
\newblock Exploring content models for multi-document summarization.
\newblock In \emph{Proceedings of Human Language Technologies: The 2009 Annual
  Conference of the North American Chapter of the Association for Computational
  Linguistics}, pages 362--370.

\bibitem[{Handler and O'Connor(2017)}]{handler2017rookie}
Abram Handler and Brendan O'Connor. 2017.
\newblock Rookie: A unique approach for exploring news archives.
\newblock In \emph{Proceedings of Data Science + Journalism workshop at KDD},
  Halifax, Nova Scotia, Canada. Association for Computing Machinery.

\bibitem[{Hendahewa and Shah(2017)}]{hendahewa2017evaluatingExpSeTrails}
Chathra Hendahewa and Chirag Shah. 2017.
\newblock Evaluating user search trails in exploratory search tasks.
\newblock \emph{Information Processing \& Management}, 53(4):905--922.

\bibitem[{Hong and Nenkova(2014)}]{hong2014wordImportance}
Kai Hong and Ani Nenkova. 2014.
\newblock \href {https://doi.org/10.3115/v1/E14-1075} {Improving the estimation
  of word importance for news multi-document summarization}.
\newblock In \emph{Proceedings of the 14th Conference of the {E}uropean Chapter
  of the Association for Computational Linguistics}, pages 712--721,
  Gothenburg, Sweden. Association for Computational Linguistics.

\bibitem[{Ingwersen(1992)}]{ingwersen1992iir}
Peter Ingwersen. 1992.
\newblock \emph{Information retrieval interaction}, volume 246.
\newblock Taylor Graham London.

\bibitem[{Kelly and Lin(2007)}]{kelly2007overviewTrec}
Diane Kelly and Jimmy Lin. 2007.
\newblock Overview of the trec 2006 ciqa task.
\newblock In \emph{ACM SIGIR Forum}, volume~41, pages 107--116. ACM New York,
  NY, USA.

\bibitem[{Leuski et~al.(2003)Leuski, Lin, and Hovy}]{leuski2003ineats}
Anton Leuski, Chin-Yew Lin, and Eduard Hovy. 2003.
\newblock \href {https://doi.org/10.3115/1075178.1075197} {i{N}e{ATS}:
  Interactive multi-document summarization}.
\newblock In \emph{The Companion Volume to the Proceedings of 41st Annual
  Meeting of the Association for Computational Linguistics}, pages 125--128,
  Sapporo, Japan. Association for Computational Linguistics.

\bibitem[{Levenshtein(1966)}]{levenshtein1966editdistance}
Vladimir~I Levenshtein. 1966.
\newblock Binary codes capable of correcting deletions, insertions, and
  reversals.
\newblock In \emph{Soviet physics doklady}, volume~10, pages 707--710.

\bibitem[{Lewis(1982)}]{lewis1982thinkaloud}
Clayton Lewis. 1982.
\newblock \emph{Using the" thinking-aloud" method in cognitive interface
  design}.
\newblock IBM TJ Watson Research Center Yorktown Heights, NY.

\bibitem[{Lewis et~al.(2013)Lewis, Utesch, and Maher}]{lewis2013umuxlite}
James~R Lewis, Brian~S Utesch, and Deborah~E Maher. 2013.
\newblock Umux-lite: when there's no time for the sus.
\newblock In \emph{Proceedings of the SIGCHI Conference on Human Factors in
  Computing Systems}, pages 2099--2102.

\bibitem[{Li et~al.(2008)Li, Wei, Lu, and He}]{li2008pnr2}
Wenjie Li, Furu Wei, Qin Lu, and Yanxiang He. 2008.
\newblock Pnr2: ranking sentences with positive and negative reinforcement for
  query-oriented update summarization.
\newblock In \emph{Proceedings of the 22nd international conference on
  computational linguistics (Coling 2008)}, pages 489--496.

\bibitem[{Lin(2004)}]{lin2004rouge}
Chin-Yew Lin. 2004.
\newblock \href {https://www.aclweb.org/anthology/W04-1013} {{ROUGE}: A package
  for automatic evaluation of summaries}.
\newblock In \emph{Text Summarization Branches Out}, pages 74--81, Barcelona,
  Spain. Association for Computational Linguistics.

\bibitem[{Lin(2007)}]{lin2007recallCurve}
Jimmy Lin. 2007.
\newblock Is question answering better than information retrieval? towards a
  task-based evaluation framework for question series.
\newblock In \emph{Human Language Technologies 2007: The Conference of the
  North American Chapter of the Association for Computational Linguistics;
  Proceedings of the Main Conference}, pages 212--219.

\bibitem[{Lin and Demner-Fushman(2006)}]{lin2006nuggetPyr}
Jimmy Lin and Dina Demner-Fushman. 2006.
\newblock Will pyramids built of nuggets topple over?
\newblock In \emph{Proceedings of the main conference on Human Language
  Technology Conference of the North American Chapter of the Association of
  Computational Linguistics}, pages 383--390. Association for Computational
  Linguistics.

\bibitem[{Lin et~al.(2010)Lin, Madnani, and Dorr}]{lin2010interactiveMMR}
Jimmy Lin, Nitin Madnani, and Bonnie Dorr. 2010.
\newblock \href {https://www.aclweb.org/anthology/N10-1041} {Putting the user
  in the loop: Interactive maximal marginal relevance for query-focused
  summarization}.
\newblock In \emph{Human Language Technologies: The 2010 Annual Conference of
  the North {A}merican Chapter of the Association for Computational
  Linguistics}, pages 305--308, Los Angeles, California. Association for
  Computational Linguistics.

\bibitem[{MacQueen et~al.(1967)}]{macqueen1967kmeans}
James MacQueen et~al. 1967.
\newblock Some methods for classification and analysis of multivariate
  observations.
\newblock In \emph{Proceedings of the fifth Berkeley symposium on mathematical
  statistics and probability}, volume~1, pages 281--297. Oakland, CA, USA.

\bibitem[{Marchionini(2006)}]{marchionini2006exploratorySearch}
Gary Marchionini. 2006.
\newblock Exploratory search: from finding to understanding.
\newblock \emph{Communications of the ACM}, 49(4):41--46.

\bibitem[{McCreadie et~al.(2014)McCreadie, Macdonald, and
  Ounis}]{mccreadie2014incremental}
Richard McCreadie, Craig Macdonald, and Iadh Ounis. 2014.
\newblock Incremental update summarization: Adaptive sentence selection based
  on prevalence and novelty.
\newblock In \emph{Proceedings of the 23rd ACM international conference on
  conference on information and knowledge management}, pages 301--310.

\bibitem[{Mihalcea and Tarau(2004)}]{mihalcea2004textrank}
Rada Mihalcea and Paul Tarau. 2004.
\newblock Textrank: Bringing order into text.
\newblock In \emph{Proceedings of the 2004 conference on empirical methods in
  natural language processing}, pages 404--411.

\bibitem[{Mikolov et~al.(2013)Mikolov, Sutskever, Chen, Corrado, and
  Dean}]{mikolov2013distributed}
Tomas Mikolov, Ilya Sutskever, Kai Chen, Greg~S Corrado, and Jeff Dean. 2013.
\newblock Distributed representations of words and phrases and their
  compositionality.
\newblock In \emph{Advances in neural information processing systems}, pages
  3111--3119.

\bibitem[{Nenkova and Passonneau(2004)}]{nenkova2004pyramid}
Ani Nenkova and Rebecca Passonneau. 2004.
\newblock \href {https://www.aclweb.org/anthology/N04-1019} {Evaluating content
  selection in summarization: The pyramid method}.
\newblock In \emph{Proceedings of the Human Language Technology Conference of
  the North {A}merican Chapter of the Association for Computational
  Linguistics: {HLT}-{NAACL} 2004}, pages 145--152, Boston, Massachusetts, USA.
  Association for Computational Linguistics.

\bibitem[{NIST(2014)}]{nist2014DUCWebsite}
NIST. 2014.
\newblock Document understanding conferences.
\newblock \url{https://duc.nist.gov/}.
\newblock Accessed: 2020-05-19.

\bibitem[{Palagi et~al.(2017)Palagi, Gandon, Giboin, and
  Troncy}]{palagi2017explSearchModels}
Emilie Palagi, Fabien Gandon, Alain Giboin, and Rapha{\"e}l Troncy. 2017.
\newblock A survey of definitions and models of exploratory search.
\newblock In \emph{Proceedings of the 2017 ACM Workshop on Exploratory Search
  and Interactive Data Analytics}, pages 3--8.

\bibitem[{Radev et~al.(2004)Radev, Jing, Sty{\'s}, and
  Tam}]{radev2004centroidMDS}
Dragomir~R Radev, Hongyan Jing, Ma{\l}gorzata Sty{\'s}, and Daniel Tam. 2004.
\newblock Centroid-based summarization of multiple documents.
\newblock \emph{Information Processing \& Management}, 40(6):919--938.

\bibitem[{Reimers and Gurevych(2019)}]{reimers2019sentenceBert}
Nils Reimers and Iryna Gurevych. 2019.
\newblock \href {https://doi.org/10.18653/v1/D19-1410} {Sentence-{BERT}:
  Sentence embeddings using {S}iamese {BERT}-networks}.
\newblock In \emph{Proceedings of the 2019 Conference on Empirical Methods in
  Natural Language Processing and the 9th International Joint Conference on
  Natural Language Processing (EMNLP-IJCNLP)}, pages 3982--3992, Hong Kong,
  China. Association for Computational Linguistics.

\bibitem[{Roit et~al.(2020)Roit, Klein, Stepanov, Mamou, Michael, Stanovsky,
  Zettlemoyer, and Dagan}]{roit2019controlled}
Paul Roit, Ayal Klein, Daniela Stepanov, Jonathan Mamou, Julian Michael,
  Gabriel Stanovsky, Luke Zettlemoyer, and Ido Dagan. 2020.
\newblock Controlled crowdsourcing for high-quality qa-srl annotation.
\newblock \emph{Proceedings of the 58th Annual Meeting of the Association for
  Computational Linguistics}.
\newblock To Appear.

\bibitem[{Rossiello et~al.(2017)Rossiello, Basile, and
  Semeraro}]{rossiello2017centroidBasedSumm}
Gaetano Rossiello, Pierpaolo Basile, and Giovanni Semeraro. 2017.
\newblock \href {https://doi.org/10.18653/v1/W17-1003} {Centroid-based text
  summarization through compositionality of word embeddings}.
\newblock In \emph{Proceedings of the {M}ulti{L}ing 2017 Workshop on
  Summarization and Summary Evaluation Across Source Types and Genres}, pages
  12--21, Valencia, Spain. Association for Computational Linguistics.

\bibitem[{Shapira et~al.(2019)Shapira, Gabay, Gao, Ronen, Pasunuru, Bansal,
  Amsterdamer, and Dagan}]{shapira2019litepyramids}
Ori Shapira, David Gabay, Yang Gao, Hadar Ronen, Ramakanth Pasunuru, Mohit
  Bansal, Yael Amsterdamer, and Ido Dagan. 2019.
\newblock \href {https://doi.org/10.18653/v1/N19-1072} {Crowdsourcing
  lightweight pyramids for manual summary evaluation}.
\newblock In \emph{Proceedings of the 2019 Conference of the North {A}merican
  Chapter of the Association for Computational Linguistics: Human Language
  Technologies, Volume 1 (Long and Short Papers)}, pages 682--687, Minneapolis,
  Minnesota. Association for Computational Linguistics.

\bibitem[{Shapira et~al.(2018)Shapira, Gabay, Ronen, Bar-Ilan, Amsterdamer,
  Nenkova, and Dagan}]{shapira2018multiLenRouge}
Ori Shapira, David Gabay, Hadar Ronen, Judit Bar-Ilan, Yael Amsterdamer, Ani
  Nenkova, and Ido Dagan. 2018.
\newblock \href {https://doi.org/10.18653/v1/D18-1087} {Evaluating multiple
  system summary lengths: A case study}.
\newblock In \emph{Proceedings of the 2018 Conference on Empirical Methods in
  Natural Language Processing}, pages 774--778, Brussels, Belgium. Association
  for Computational Linguistics.

\bibitem[{Shapira et~al.(2017)Shapira, Ronen, Adler, Amsterdamer, Bar-Ilan, and
  Dagan}]{shapira2017ias}
Ori Shapira, Hadar Ronen, Meni Adler, Yael Amsterdamer, Judit Bar-Ilan, and Ido
  Dagan. 2017.
\newblock \href {https://doi.org/10.18653/v1/D17-2019} {Interactive abstractive
  summarization for event news tweets}.
\newblock In \emph{Proceedings of the 2017 Conference on Empirical Methods in
  Natural Language Processing: System Demonstrations}, pages 109--114,
  Copenhagen, Denmark. Association for Computational Linguistics.

\bibitem[{Wang and Li(2010)}]{wang2010updateSumm}
Dingding Wang and Tao Li. 2010.
\newblock Document update summarization using incremental hierarchical
  clustering.
\newblock In \emph{Proceedings of the 19th ACM international conference on
  Information and knowledge management}, pages 279--288.

\bibitem[{Wessa(2020)}]{Wessa2020BootstrapTool}
Patrick Wessa. 2020.
\newblock Free statistics software, office for research development and
  education, version 1.2.1.
\newblock \url{https://www.wessa.net/}.
\newblock Accessed: 2020-05-19.

\bibitem[{White et~al.(2008)White, Marchionini, and
  Muresan}]{white2008evaluatingExpS}
Ryen~W White, Gary Marchionini, and Gheorghe Muresan. 2008.
\newblock Evaluating exploratory search systems.
\newblock \emph{Information Processing and Management}, 44(2):433.

\bibitem[{White and Roth(2009)}]{white2009exploratory}
Ryen~W White and Resa~A Roth. 2009.
\newblock \emph{Exploratory search: Beyond the query-response paradigm}.
\newblock 3. Morgan \& Claypool Publishers.

\bibitem[{Wold et~al.(1987)Wold, Esbensen, and Geladi}]{wold1987pca}
Svante Wold, Kim Esbensen, and Paul Geladi. 1987.
\newblock Principal component analysis.
\newblock \emph{Chemometrics and intelligent laboratory systems},
  2(1-3):37--52.

\bibitem[{Yan et~al.(2011)Yan, Nie, and Li}]{yan2011personalizedSumm}
Rui Yan, Jian-Yun Nie, and Xiaoming Li. 2011.
\newblock \href {https://www.aclweb.org/anthology/D11-1124} {Summarize what you
  are interested in: An optimization framework for interactive personalized
  summarization}.
\newblock In \emph{Proceedings of the 2011 Conference on Empirical Methods in
  Natural Language Processing}, pages 1342--1351, Edinburgh, Scotland, UK.
  Association for Computational Linguistics.

\bibitem[{Yin and Pei(2015)}]{yin2015neuralMDS}
Wenpeng Yin and Yulong Pei. 2015.
\newblock Optimizing sentence modeling and selection for document
  summarization.
\newblock In \emph{Twenty-Fourth International Joint Conference on Artificial
  Intelligence}.

\bibitem[{Zhang* et~al.(2020)Zhang*, Kishore*, Wu*, Weinberger, and
  Artzi}]{zhang2019bertscore}
Tianyi Zhang*, Varsha Kishore*, Felix Wu*, Kilian~Q. Weinberger, and Yoav
  Artzi. 2020.
\newblock \href {https://openreview.net/forum?id=SkeHuCVFDr} {Bertscore:
  Evaluating text generation with bert}.
\newblock In \emph{International Conference on Learning Representations}.

\bibitem[{Zhao et~al.(2009)Zhao, Wu, and Huang}]{zhao2009qfs}
Lin Zhao, Lide Wu, and Xuanjing Huang. 2009.
\newblock Using query expansion in graph-based approach for query-focused
  multi-document summarization.
\newblock \emph{Information processing \& management}, 45(1):35--41.

\bibitem[{Zopf et~al.(2016)Zopf, Menc{\'\i}a, and
  F{\"u}rnkranz}]{zopf2016sequentialUpdateSumm}
Markus Zopf, Eneldo~Loza Menc{\'\i}a, and Johannes F{\"u}rnkranz. 2016.
\newblock Sequential clustering and contextual importance measures for
  incremental update summarization.
\newblock In \emph{Proceedings of COLING 2016, the 26th International
  Conference on Computational Linguistics: Technical Papers}, pages 1071--1082.

\bibitem[{Zuccon et~al.(2013)Zuccon, Leelanupab, Whiting, Yilmaz, Jose, and
  Azzopardi}]{zuccon2013crowdsourcingInteractions}
Guido Zuccon, Teerapong Leelanupab, Stewart Whiting, Emine Yilmaz, Joemon~M
  Jose, and Leif Azzopardi. 2013.
\newblock Crowdsourcing interactions: using crowdsourcing for evaluating
  interactive information retrieval systems.
\newblock \emph{Information retrieval}, 16(2):267--305.

\end{thebibliography}
\bibliographystyle{acl_natbib}

\clearpage
\appendix
These \textbf{appendices} provide further details on some issues within the main paper, mostly technicalities, as well as input and output examples.

\section{Implementation Details}
\paragraph{BERT-based implementations.}
While developing the initial summary and query-response algorithms, we also experimented with BERT variants.

In $I^{\mathrm{CL}}$, the sentence-averaged Word2Vec embeddings were replaced with Sentence-BERT \citep{reimers2019sentenceBert} representations, drawing on the `roberta-base-nli-stsb-mean-tokens' model.\footnote{\url{https://github.com/UKPLab/sentence-transformers}}

As an alternate method for computing similarity between a query and sentences, we used BERTScore \citep{zhang2019bertscore}.

Both were time-expensive and did not substantially improve, if at all, the outputs.

\paragraph{Initial summary hyper-parameters.}
As mentioned in the paper, the $I^{\mathrm{CL}}$ hyper-parameters were lightly adjusted by computing ROUGE scores of some outputs against reference summaries and ensuring fast processing. For sentence-representative vector dimension (final chosen value of 20) we tested several values between 10 and 100. For number of sentence clusters (30) we tested values between 10 and 50, and for similarity threshold (0.95) we tested several options within the 0 to 1 range.

\paragraph{Initial summary implementations.}
The sentence representations based averaged word2vec embeddings are implemented with the SpaCy library.\footnote{\url{https://spacy.io/usage/vectors-similarity}} PCA and k-means are implemented with sklearn.\footnote{\url{https://scikit-learn.org}}

The TextRank algorithm is coded with the pytextrank pipeline component in the SpaCy library.\footnote{\url{https://spacy.io/universe/project/spacy-pytextrank}}

\paragraph{Suggested queries.}
An additional element in preparing the $\mathrm{Sug}^{\mathrm{FREQ}}$ list is that an n-gram with a Levenshtein distance \citep{levenshtein1966editdistance} of less than 2 from an already selected n-gram is skipped.

\paragraph{Web application.}
A previous version of the web application included a button for \textit{additional general information}, which was supported by QFSE System $S_1$. This sent an empty query to the system, for which the sentence from the next unused cluster in the $I^{\mathrm{CL}}$ algorithm was returned (rotating to the first cluster if all clusters have been used, and taking the next best unused representing sentence). In our preliminary user study and crowdsourcing experiments, we found that this feature was mainly a distraction and induced exploration laziness.

The Web application is implemented in HTML, CSS and Javascript, and the backend in Python. The app communicates with the backend over standard HTTP Post requests in JSON format.

\paragraph{Server specifications.}
We ran experiments, and run our QFSE systems on an Intel Xeon CPU E5-2670 v3 @ 2.30GHz server with 50GB RAM. Run times are similar on an Intel Core i7-6600 CPU @ 2.60GHz laptop with 16GB RAM.

To check if the query-response component with BERTScore would be more practical on a GPU server, we tested it on an Nvidia TITAN X GPU server. A single query response took 40-50 seconds to compute.

\section{Controlled Crowdsourcing Details}
The controlled crowdsourcing protocol finds high quality users for the collection of system sessions. The \textit{use-case} we enforced was producing an informative summary draft text which a journalist could use to best produce an overview of the given topic for the general public.

\subsection{Trap Task}
This task, consisting of three questions, aims to discover workers with an ability to apprehend salient information within text. It was implemented standardly within the Amazon Mechanical Turk\footnote{\url{https://www.mturk.com}} platform.

\paragraph{Question 1.}
This question tests apprehension of the notion of a general summary, by asking the user to choose a sentence that would be best to include in an overview of some topic, given the topic name and four relevant sentences of varying informativeness.

For our journalistic scenario, we randomly selected 10 topics from the DUC\footnote{\url{https://duc.nist.gov/}} 2007 MDS dataset that include Pyramid SCUs \citep{nenkova2004pyramid}, and for each topic manually chose one SCU with a weight of 4, and three SCUs with a weight of 1. The SCU with the higher weight would be expected to be the more appropriate choice for the journalistic overview, since all four reference summaries of the topic include it.

\paragraph{Question 2.}
This question simulates a scenario closer to interactive summarization. A three-sentence ``initial summary'' of the same topic (as Question 1) is presented, and the worker is asked to choose the best of three possible interaction possibilities that would provide more information on the topic from a theoretical search tool. We presented three pairs of queries of varying relevance and informativeness as interaction possibilities.

The ``initial summary'' is the lead-three sentences of one of the reference summaries of the topic. We prepared the three choices of two queries based on salient phrases manually found within the reference summaries. One pair of queries is worthy, one pair is somewhat worthy, and a third pair is unworthy.

\paragraph{Question 3.}
This question tests attentiveness to the task and creativity by asking users to suggest another interaction (query) to the theoretical search tool. Such an open-ended question requires more thought and hence filters careless or guessing workers.

\paragraph{Task preparation.}
We ran the 10 tasks internally with research colleagues to find vulnerabilities, and edited some questions accordingly. In addition we recorded the average work time to set a fair task payment on the crowdsourcing platform. In retrospect, the time it took crowd-workers was about two-thirds of the time it took internal workers. Moreover, the time was an additional indication of better workers -- those completing the task correctly in shorter time were likely superior.

We paid \$0.50 for each trap task assignment, estimating about 3.5 minutes of work time. Good workers completed the task in 2.5 minutes on average, which should fairly pay only \$0.30.

\paragraph{Task assessment.}
The first two questions are automatic filters for insincere workers. A meaningful answer to the third question, assessed manually, serves as a sanity check which we found useful for additional filtering.

The workers passing this phase were contacted via email. The message included an explanation and estimated payment of the subsequent tasks.

\subsection{Practice Task}
This task is an external question done within an IFrame on Mechanical Turk.

Two practice tasks were prepared from DUC 2006, separate from the 20 used for real session collection. Workers completing both tasks with predominantly relevant queries (checked manually) were asked to continue on to the final task.

We emphasized the use case of preparing a journalistic overview by instructing to ``produce an informative summary draft text which a journalist could use to best produce an overview of the topic''.

We paid \$0.90 for each practice task, estimating 6-7 minutes of work per assignment. Our estimate was about right.

\subsection{Evaluation Session Collection Task}
As before, this external question task is done within an IFrame on Mechanical Turk.

For the session collection tasks we paid \$0.70 per topic, estimating 5 minutes of work. We promised to give a bonus for good work, to motivate completion of more assignments, and in higher standards. We awarded \$0.15 to \$0.30 bonus according to the quality (assessed manually), per assignment. All sessions were of very high quality, but some made an extra effort and provided comments and feedback.

\subsection{Wording of Human Ratings}
For our journalistic use-case, the ratings within a session are worded as follows:
\begin{itemize}
    \item $[R.1]$ ``How useful is this for the journalist's generic overview of the topic?''
    \item $[R.2]$ ``How much useful info does this add to the journalist's overview (regardless of how well it matched your query)?''
    \item $[R.3]$ ``During the interactive stage, how well did the responses respond to your queries?''
    \item $[R.4]$ ``As a system for exploring information on a topic,
    \begin{itemize}
        \item $[R.4a]$ ``its capabilities meet the need to efficiently collect useful information for a journalistic overview.''
        \item $[R.4b]$ ``it is easy to use.''
    \end{itemize}
\end{itemize}

\subsection{Wild Crowdsourcing}
For quality control, at the end of a session the user filled a questionnaire, in which they mark whether 10 statements are covered in their generated session. Of those statements, five were (separately) crowdsourced summary content units (SCUs) from the topic's reference summaries \citep{shapira2019litepyramids}, one of those SCUs was repeated to test for identical markings, two statements were SCUs from another topic, and two statements were the two shortest sentences output by the session. We thus know the answers to 4 statements and have a repeating statement test. Sessions with minimal mistakes could hypothetically be considered sincere.

\section{Experiments}
\subsection{Data}
Systems were evaluated using data from the DUC 2006 MDS dataset. 20 topics were used (all those with Pyramid \citep{nenkova2004pyramid} evaluations). These are: D0601, D0603, D0605, D0608, D0614, D0615, D0616, D0617, D0620, D0624, D0627, D0628, D0629, D0630, D0631, D0640, D0643, D0645, D0647, D0650. The practice tasks in the controlled crowdsourcing procedure used topics D0602, D0606. The 10 topics in the trap task are based on document sets with Pyramid evaluations from DUC 2007. These are: D0701A, D0703A, D0704A, D0705A, D0706B, D0707B, D0710C, D0711C, D0714D, D0716D.

\subsection{More Results}
Similar to the results on ROUGE-1 in the main paper, Tables \ref{table_scoresR2}, \ref{table_scoresRL} and \ref{table_scoresRSU} are for metrics ROUGE-2, ROUGE-L and ROUGE-SU respectively.
\begin{table}[h]
    \centering
    \resizebox{\columnwidth}{!}{
    \begin{tabular}{l|c|c|c}
        \hline
        \textbf{Sessions} & \textbf{S@L 150} & \textbf{S@L 250} & \textbf{S@L 350} \\
        \hline
        $S_1$ $L^{Oracle}$ & $.063$ ($\pm .011$) & $.085$ ($\pm .013$) & $.096$ ($\pm .013$) \\
        $S_1$ Real & $.064$ ($\pm .010$) & $.077$ ($\pm .010$) & $.082$ ($\pm .010$) \\
        $S_1$ $L^{Sug}$ & $.065$ ($\pm .009$) & $.078$ ($\pm .010$) & $.082$ ($\pm .012$) \\
        \hline
        $S_2$ $L^{Oracle}$ & $.067$ ($\pm .011$) & $.085$ ($\pm .013$) & $.094$ ($\pm .013$) \\
        $S_2$ Real & $.058$ ($\pm .010$) & $.072$ ($\pm .011$) & $.077$ ($\pm .013$) \\
        $S_2$ $L^{Sug}$ & $.056$ ($\pm .010$) & $.068$ ($\pm .011$) & $.073$ ($\pm .011$) \\
        \hline
    \end{tabular}}
    \caption{ROUGE-2 $F_1$-based average scores of simulated sessions vs.\ controlled crowdsourced sessions. Scores at 350 words are approximate as a few sessions were shorter. Intervals at $\geq 95\%$ confidence.}
    \label{table_scoresR2}
\end{table}

\begin{table}[h]
    \centering
    \resizebox{\columnwidth}{!}{
    \begin{tabular}{l|c|c|c}
        \hline
        \textbf{Sessions} & \textbf{S@L 150} & \textbf{S@L 250} & \textbf{S@L 350} \\
        \hline
        $S_1$ $L^{Oracle}$ & $.270$ ($\pm .013$) & $.328$ ($\pm .012$) & $.333$ ($\pm .014$) \\
        $S_1$ Real & $.271$ ($\pm .010$) & $.314$ ($\pm .010$) & $.319$ ($\pm .010$) \\
        $S_1$ $L^{Sug}$ & $.258$ ($\pm .010$) & $.299$ ($\pm .011$) & $.302$ ($\pm .011$) \\
        \hline
        $S_2$ $L^{Oracle}$ & $.275$ ($\pm .012$) & $.327$ ($\pm .015$) & $.332$ ($\pm .015$) \\
        $S_2$ Real & $.270$ ($\pm .011$) & $.313$ ($\pm .014$) & $.315$ ($\pm .014$) \\
        $S_2$ $L^{Sug}$ & $.271$ ($\pm .011$) & $.311$ ($\pm .014$) & $.313$ ($\pm .013$) \\
        \hline
    \end{tabular}}
    \caption{ROUGE-L $F_1$-based average scores of simulated sessions vs.\ controlled crowdsourced sessions. Scores at 350 words are approximate as a few sessions were shorter. Intervals at $\geq 95\%$ confidence.}
    \label{table_scoresRL}
\end{table}

\begin{table}[h]
    \centering
    \resizebox{\columnwidth}{!}{
    \begin{tabular}{l|c|c|c}
        \hline
        \textbf{Sessions} & \textbf{S@L 150} & \textbf{S@L 250} & \textbf{S@L 350} \\
        \hline
        $S_1$ $L^{Oracle}$ & $.091$ ($\pm .008$) & $.145$ ($\pm .008$) & $.156$ ($\pm .011$) \\
        $S_1$ Real & $.090$ ($\pm .007$) & $.137$ ($\pm .009$) & $.145$ ($\pm .009$) \\
        $S_1$ $L^{Sug}$ & $.089$ ($\pm .007$) & $.133$ ($\pm .009$) & $.139$ ($\pm .008$) \\
        \hline
        $S_2$ $L^{Oracle}$ & $.093$ ($\pm .008$) & $.145$ ($\pm .010$) & $.156$ ($\pm .013$) \\
        $S_2$ Real & $.090$ ($\pm .006$) & $.137$ ($\pm .010$) & $.141$ ($\pm .011$) \\
        $S_2$ $L^{Sug}$ & $.090$ ($\pm .007$) & $.133$ ($\pm .012$) & $.140$ ($\pm .011$) \\
        \hline
    \end{tabular}}
    \caption{ROUGE-SU $F_1$-based average scores of simulated sessions vs.\ controlled crowdsourced sessions. Scores at 350 words are approximate as a few sessions were shorter. Intervals at $\geq 95\%$ confidence.}
    \label{table_scoresRSU}
\end{table}

\subsection{Length@Score Metric}
\begin{table}[h]
    \centering
    \resizebox{\columnwidth}{!}{
    \begin{tabular}{l|c|c|c|c}
        \hline
        \textbf{Sessions} & \textbf{R1 .37} & \textbf{R2 .075} & \textbf{RL .31} & \textbf{RSU .14}  \\
        \hline
        $S_1$ $L^{Oracle}$ & 193 & 191 & 199 & 232 \\
        $S_1$ Real & 218 & 233 & 233 & 266 \\
        $S_1$ $L^{Sug}$ & 231 & 200 & 253 & N/A \\
        \hline
        $S_2$ $L^{Oracle}$ & 192 & 180 & 197 & 232 \\
        $S_2$ Real & 221 & 288 & 237 & 269 \\
        $S_2$ $L^{Sug}$ & 236 & N/A & 245 & 310 \\
        \hline
    \end{tabular}}
    \caption{The Length@Score measurements for ROUGE-1, ROUGE-2, ROUGE-L and ROUGE-SU $F_1$ scores. This answers how many words on average are needed to reach the specified ROUGE $F_1$ score. Values are calculated from the averaged overall session of a system, and not as a macro-average. When a value is `N/A', the system did not reach the score.}
    \label{table_lengthAtScore}
\end{table}
We also computed a ``Length@Score'' measurement assessing at what word length, on average, a given content score can be reached by the system. It might forecast how much interaction is required to reach a certain information coverage, indicating system effectiveness. This is computed on the overall averaged session, and not as a macro-average, since some topics do not reach the specified score. By looking at the averaged session, we have a better overlook at the system's capability. Table \ref{table_lengthAtScore} presents these values. Here, the ROUGE scores for which to compute the resulting lengths are chosen based on numbers within a range of scores found in MDS literature employing the DUC 2006 dataset on extractive summarization systems. We see an overall similar trend where the controlled user sessions fall between the upper and lower bounds, and that $S_1$ scores slightly better than $S_2$. As an analysis, we can compare, e.g., the upper and lower bounds on ROUGE-1, and observe that the lower bound requires about 40 words more (20\%) to contain a similar amount of salient content.

\subsection{Standard Metric Implementations}
\paragraph{Confidence intervals.}
All confidence intervals were calculated with a Python bootstrapping library,\footnote{\url{https://pypi.org/project/bootstrapped/}} and sometimes validated with an online tool \citep{Wessa2020BootstrapTool}. The confidence level is $\geq95\%$ throughout the paper.

\paragraph{ROUGE.}
ROUGE scores were obtained either with the rouge Python package\footnote{\url{https://pypi.org/project/rouge/}} for quick sentence comparisons, or the pyrouge Python package\footnote{\url{https://pypi.org/project/pyrouge/}} for file comparison (long text to reference summaries).

\paragraph{UMUX-Lite}
When on a 5-point scale, the UMUX-Lite \citep{lewis2013umuxlite} score is computed:
\[0.65*(([R.4a]+[R.4b]-2)*(100/8))+22.9\]
68 is considered average, and above 80 is excellent.

\subsection{Real Session Examples}
\begin{table*}
    \normalsize
    \centering
    \resizebox{\linewidth}{!}{
    \begin{tabular}{|p{2cm}|p{15cm}|}
        \hline
        \textbf{Query} & \textbf{Response} \\ \hline
        [I] & ``The weather pattern comes on the heels of an El Nino that dissipated in 1998. El Nino is a natural phenomenon that involves a surface warming of the eastern and central Pacific Ocean around the equator. For the United States, an El Nino brings unusually warm temperatures to the northern states and cooler, wetter conditions to the South. By the time the next El Nino arrives, scientists hope to be able to give much better forecasts.'' \\ \hline
        ``El Nino'' [H] & ``La Ninas sometimes occur between El Ninos. Will La Nina wane as El Nino lingers?'' he said.'' \\ \hline
        ``Pacific Ocean'' [H] & ``Those conditions are warming in the eastern Pacific Ocean; shifting of precipitation from the western to the eastern Pacific; slowing of trade winds; and changes in sea level. El Ninos increase storm activity in California and South America while calming things down in the Atlantic and Caribbean.'' \\ \hline
        ``equator'' [H] & ``Scientists at NASA's Jet Propulsion Laboratory came the conclusion after analyzing the data collected from satellite last week. The weather-altering La Nina phenomenon has all but vanished from the Pacific Ocean, according to NASA satellite observations.'' \\ \hline
        ``temperatures'' [H] & ``Sea surface temperatures throughout the equatorial east-central Pacific increased during April and May, when temperatures normally decrease in this region. He said the center predicted that after rapid cooling during the summer, ocean temperatures would rebound before cooling again.'' \\ \hline
        ``forecast'' [H] & ``It disrupts weather worldwide, causing storms, floods and droughts. It is blamed for flooding and heavy snow in the West, abnormally warm weather in the East and extreme weather from South America to Asia.'' \\ \hline
        ``la nina'' [S] & ``La Nina and El Nino form the opposite ends of the same climatic cycle, with El Nino fueled by unusually warm water in the eastern Pacific and La Nina driven by cold. A La Nina can drop it nearly 8 degrees.'' \\ \hline
        ``weather patterns'' [S] & ``El Nino causes reversals of normal weather patterns, resulting in drought in usually wet locales and flooding in arid areas. He said that soil moisture and snow cover appear perceptibly to influence seasonal weather patterns, adding that computer models at Scripps are beginning to incorporate some of these variables with some success.'' \\ \hline
        ``equatorial pacific'' [S] & ``Wetter conditions than usual will continue over the central and eastern equatorial Pacific and over southeastern South America. La Nina refers to the phenomenon of rapidly cooling equatorial waters in the central Pacific.'' \\ \hline
        ``weather phenomenon'' [S] & ``El Nino phenomenon, which comes at an interval of roughly four years, refers to abnormal temperature rises in tropical waters which normally leads to global anomalies in weather and climate. El Nino is a periodic warming of the water in the eastern Pacific Ocean, which shifts weather patterns arriving from that area and can affect the climate worldwide.'' \\ \hline
        ``surface temperatures'' [S] & ``During August and September, ocean surface temperatures reached near-record levels in many sections of the equatorial Pacific. The pool of unusually cool tropical water marked by low sea levels has shriveled and the equatorial Pacific is warming to normal temperatures, NASA's Jet Propulsion Laboratory said this week.'' \\ \hline
        ``global warming'' [S] & ``Despite causing anomaly of weather and climate, the El Nino event may help cut global warming by temporarily stemming the release of carbon dioxide from the Pacific Ocean, scientists suggested Thursday. The program said experiments on the relationship between increasing carbon dioxide levels and global warming being done at the Mauna Loa Observatory in Hawaii supported the Australian claims.'' \\ \hline
        ``carbon dioxide'' [S] & ``The oceans exchange large amounts of carbon dioxide with the atmosphere each year. Scientists usually estimate how much carbon dioxide should be in the atmosphere by calculating how much fossil fuel is burned." \\
        \hline
    \end{tabular}}
    \caption{A real session in System $S_1$ on the topic ``El Ni{\~n}o''. [I] stands for `initial summary', [H] stands for `highlight' and [S] stands for `suggested query'. The user here starts out with highlighted queries, and then uses suggestions.}
    \label{chart_sessionS1}
\end{table*}
\begin{table*}
    \normalsize
    \centering
    \resizebox{\linewidth}{!}{
    \begin{tabular}{|p{1.5cm}|p{17.5cm}|}
        \hline
        \textbf{Query} & \textbf{Response} \\ \hline
        [I] & ``The good news is that they can now forecast an El Nino with some precision, and during El Nino years predict its effect on the world's weather months in advance. The conclusion was made by Song Jiaxi and his colleagues with the National Marine Environmental Forecasting Center in their annual marine disaster forecast report, which was released today. Despite causing anomaly of weather and climate, the El Nino event may help cut global warming by temporarily stemming the release of carbon dioxide from the Pacific Ocean, scientists suggested Thursday.'' \\ \hline
        ``climate change'' [F] & ``In general, a large proportion of infectious disease agents are very sensitive to slight changes in climate,\" said McMichael, who was not involved in the research. Conditions like El Nino might settle in almost permanently if global warming gets bad enough, making climate disruptions such as droughts or excessive winter rain essentially the norm, a computer study suggests.'' \\ \hline
        [R] & ``El Nino is a periodic warming of the water in the eastern Pacific Ocean, which shifts weather patterns arriving from that area and can affect the climate worldwide. Scientists cautioned that like its warm counterpart, El Nino, a La Nina condition will influence global climate and weather until it has completely subsided.'' \\ \hline
        [R] & ``WMO added that uncertainty over surface temperatures in the Atlantic and Indian Oceans, which contribute to changing rainfall over Africa and South America, makes an accurate forecast difficult. El Nino phenomenon, which comes at an interval of roughly four years, refers to abnormal temperature rises in tropical waters which normally leads to global anomalies in weather and climate.'' \\ \hline
        ``countries affected'' [F] & ``Diarrhea kills as many as 3 million children under the age of 5 worldwide every year and sickens millions more, mostly in developing countries. The phenomenon had been responsible for only 40 percent rainfall in the country in June, he said.'' \\ \hline
        [R] & ``When the present levels of the greenhouse gas carbon dioxide were doubled in the experiment, the number of El Ninos affecting Australia nearly doubled too, the scientist said. La Ninas, by contrast, reduce storms in California but stir up trouble in other parts of the country as well as in India and southeast Asia.'' \\ \hline
        ``la nina'' [F] & ``La Ninas sometimes occur between El Ninos. Will La Nina wane as El Nino lingers?'' he said.'' \\ \hline
        [R] & ``La Nina and El Nino form the opposite ends of the same climatic cycle, with El Nino fueled by unusually warm water in the eastern Pacific and La Nina driven by cold. A La Nina can drop it nearly 8 degrees.'' \\ \hline
        [R] & ``If La Nina dissipates before it hits Los Angeles, the area could face a more typical wet winter. La Nina, Spanish for ``little girl,'' is just the opposite, with the warm conditions of El Nino returning to the west.'' \\ \hline
        ``global warming'' [F] & ``The program said experiments on the relationship between increasing carbon dioxide levels and global warming being done at the Mauna Loa Observatory in Hawaii supported the Australian claims. Australian scientists have uncovered a link between global warming and the increasing frequency of the El Nino weather system, the Australian Broadcasting Corporation (ABC) reported tonight.'' \\ \hline
        [R] & ``Gerald Meehl of the National Center for Atmospheric Research in Boulder, Colo., agreed it will take more study to understand how global warming might affect El Nino. The potential effect of global warming on disease is controversial,\" said William Checkley of Baltimore-based Johns Hopkins, who led the study.'' \\ \hline
        [R] & ``The world was also warmer in that quiet El Nino period, further pointing to a possible relationship between El Nino and global warming. The computer simulation suggests that unrestrained global warming could set up the same kind of pattern in ocean surface temperatures, but as a more or less constant condition.'' \\ \hline
        [R] & ``The Four Corners TV program claimed scientific data were showing that a man-made permanent drought was in the offing as a result of global warming. El Nino is the name given to the unusual warming of waters in the Pacific Ocean that affects global wind and temperature patterns.'' \\ \hline
        [R] & ``The pool of unusually cool tropical water marked by low sea levels has shriveled and the equatorial Pacific is warming to normal temperatures, NASA's Jet Propulsion Laboratory said this week. El Nino is a natural phenomenon that involves a surface warming of the eastern and central Pacific Ocean around the equator.'' \\ \hline
        [R] & ``El Nino is the name given to the unusual warming of waters in the Pacific Ocean that affects wind and temperature patterns. BEIJING, May 18 (Xinhua) -- Chinese oceanographers today expect climatic repercussions on global weather as El Nino is expected to come sometime this year or the next.'' \\ \hline
        ``oceans'' [F] & ``It is estimated that the equatorial Pacific Ocean is the source of 72 percent of the annual flux of carbon to the atmosphere from the equatorial oceans. Those conditions are warming in the eastern Pacific Ocean; shifting of precipitation from the western to the eastern Pacific; slowing of trade winds; and changes in sea level.'' \\ \hline
        [R] & ``The oceans exchange large amounts of carbon dioxide with the atmosphere each year. During August and September, ocean surface temperatures reached near-record levels in many sections of the equatorial Pacific.'' \\ \hline
        [R] & ``The most important of gas-releasing areas is the equatorial Pacific, where is also the home of El Nino, the quasi-biennial warming event in the surface waters of the eastern equatorial Pacific Ocean. Oklahoma's catastrophic tornadoes were influenced by La Nina, the weather phenomenon 5,000 miles west of the prairie in the Pacific Ocean, scientists say." \\ \hline
    \end{tabular}}
    \caption{A real session in System $S_2$ on the topic ``El Ni{\~n}o''. [I] stands for `initial summary', [F] stands for `free-text' and [R] stands for `repeat last query'. The user's strategy here seems to be repeating a query until enough information is covered on the matter.}
    \label{chart_sessionS2}
\end{table*}
Tables \ref{chart_sessionS1} and \ref{chart_sessionS2} present real sessions on Systems $S_1$ and $S_2$ respectively, both on the topic ``El Ni{\~n}o''. The two users here have different querying strategies for acquiring information. Their strategies are not consistent across topics, as they depend on the complexity and breadth of the topics.

\subsection{Simulation Samples}
\begin{table*}
    \normalsize
    \centering
    \resizebox{\linewidth}{!}{
    \begin{tabular}{|p{10cm}|p{10cm}|}
        \hline
        \multicolumn{2}{|c|}{$L^{\mathrm{Oracle}}$} \\
        \hline
        \multicolumn{2}{|l|}{[I]} \\
        \multicolumn{2}{|l|}{``El Nino can cause storms in California, tornadoes in Florida, a mild winter in the northern states.''} \\
        \multicolumn{2}{|l|}{``Scientists in the United States, Australia, Israel, and Germany are using sophisticated computer simulations.''} \\
        \multicolumn{2}{|l|}{``La Nina works in reverse of El Nino.''} \\
        \multicolumn{2}{|l|}{``La Nina is the phenomenon of rapidly cooling equatorial waters in the central Pacific.''} \\
        \multicolumn{2}{|l|}{``Computer modeling a simulation are used to study El Nino and El Nina patterns.''} \\
        \multicolumn{2}{|l|}{``El Nino typically lasts a year.''} \\
        \multicolumn{2}{|l|}{``Scientific technologies and techniques for studying these phenomena include computer modeling.''} \\
        \multicolumn{2}{|l|}{``El Nino may lessen global warming by temporarily stemming the release of carbon dioxide from the Pacific.''} \\
        \multicolumn{2}{|l|}{``Computer module studies and satellite systems allow for a better understanding of how El Nino and La Nina form.''} \\
        \multicolumn{2}{|l|}{``The results of El Nino and El Nina can have severe economic impacts, disease and death.''} \\
        \hline
        \multicolumn{2}{|c|}{$L^{\mathrm{Sug}}$} \\
        \multicolumn{1}{|c}{$\mathrm{Sug}^{\mathrm{FREQ}}$} & \multicolumn{1}{c|}{$\mathrm{Sug}^{\mathrm{TR}}$} \\
        \hline
        [I] & [I] \\
        ``el nino'' & ``el nino years'' \\
        ``la nina'' & ``el nino phenomenon'' \\
        ``pacific ocean'' & ``el nino events'' \\
        ``carbon dioxide'' & ``el nino activity'' \\
        ``weather patterns'' & ``next el nino'' \\
        ``equatorial pacific'' & ``el nino behavior'' \\
        ``south america'' & ``el nino update'' \\
        ``global warming'' & ``el ninos'' \\
        ``weather phenomenon'' & ``global weather'' \\
        ``surface temperatures'' & ``normal weather patterns'' \\
        \hline
    \end{tabular}}
    \caption{The query lists for the topic ``El Ni{\~n}o'' used in the simulations for the upper ($L^{\mathrm{Oracle}}$) and lower ($L^{\mathrm{Sug}}$) bounds. $\mathrm{Sug}^{\mathrm{FREQ}}$ is used in System $S_1$ and $\mathrm{Sug}^{\mathrm{TR}}$ is used in System $S_2$. [I] stands for ``initial summary''.}
    \label{chart_suggQueries}
\end{table*}
Table \ref{chart_suggQueries} shows the lists of queries in the simulations used for upper and lower bounds, for the topic ``El Ni{\~n}o''.

\subsection{Feedback from Users}
\begin{table*}
    \centering
    \begin{tabular}{|c|p{3cm}|p{11cm}|}
        \hline
         & \textbf{Topic} & \textbf{Comment}  \\
        \hline
        $S_1$ & School Safety & ``A...mix of information...from...statistics...to facts about... specific incidents, probably because it's such a large topic.'' \\ \hline
        $S_1$ & EgyptAir Crash & ``Searching for ``time'' didn't give any kind of date or actual time of crash, however, searching for ``date'' tended to give actual date and time together...'' \\ \hline
        $S_1$ & EgyptAir Crash & ``I noticed the search engine returned flexible dates this time (I searched 1991 and got 1996 results, for example) and I really appreciated that.'' \\ \hline
        $S_1$ & Osteoarthritis & ``There wasn't much...when trying to find specifics like symptoms...``Treatment'' pulled up the closest and most relevant results, but the others went into the weeds or pulled up things that were tangential to the search terms.'' \\ \hline
        $S_1$ & Evolution Teaching & ``...This is the first of these where I thought the system really did not meet the need to efficiently collect useful info for a journalistic overview.'' \\ \hline
        $S_1$ & Stephen Lawrence Killing & ``I was very satisfied with the information provided...a topic with which I was totally unfamiliar. More generally...the system consistently did quite well...if I were a journalist writing overviews of these topics...I would be very pleased with...the information provided by the system [and] its ease of use!...'' \\ \hline
        $S_1$ & Quebec Separatist Movement & ``I think anything that requires a higher level of background knowledge is a lot harder to research with this system, since you only get snippets.'' \\ \hline
        \hline
        $S_2$ & Wetlands & ``I was able to find information on many different aspects of the topic.'' \\ \hline
        $S_2$ & EgyptAir Crash & "...I [tried to] find out if there were other crashes...which did not turn up any info, but then later found that information when looking up a different search term." \\ \hline
        $S_2$ & Concorde Aircraft & ``The search results don't always seem to correspond to the terms keyed in...'' \\ \hline
        $S_2$ & Concorde Aircraft & ``It seems like I'm always looking for more general info...Things I would want included in an overview or even in an article dealing with a specific incident such as in this case...'' \\ \hline
        $S_2$ & Elian Gonzales & ``Most of the responses matched pretty well with the keyword search..." \\ \hline
        $S_2$ & Elian Gonzales & ``I find that I'm using the system the same way I use Google; whatever I'm wondering about I just ask in the form of a question.'' \\ \hline
        $S_2$ & US Affordable Housing & ``This set was very responsive and got results that I had not expected..." \\ \hline
        $S_2$ & Kursk Submarine & ``Outstanding! I feel like I could write an overview of this right now!'' \\ \hline
        $S_2$ & Jimmy Carter International & ``In this one, the topic...was so general that it took me a bit to figure out exactly what I was supposed to be looking for. Once I got it, everything worked fine!'' \\ \hline
        $S_2$ & El Ni{\~n}o & ``Great! Tons of useful information for a journalistic overview!'' \\ \hline
    \end{tabular}%}
    \caption{Some of many comments provided by the controlled crowdsourcing users for the two systems $S_1$ and $S_2$ on different topics (some shortened for brevity). The comments indicate that users follow the use-case. Notice that some comments show the need for prolonged exploration and human assistance for finding salient information.}
    \label{table_comments}
\end{table*}
The session collection task also had a comment box to send any kind of feedback. Table \ref{table_comments} shows a few of these comments (some shortened for brevity). The comments strongly emphasize the users' sincerity in following the use-case, and note usefulness of the system as well as provide ideas for improvements. Some comments attest to the need for prolonged exploration and human assistance for finding salient information.

\end{document}